\documentclass[10pt,twocolumn,letterpaper]{article}

\usepackage{cvpr}

\newcommand{\dinoalign}{\textsc{Dino}\textsubscript{align}}
\newcommand{\clipsim}{\textsc{Clip}\textsubscript{sim}}

\definecolor{cvprblue}{rgb}{0.21,0.49,0.74}
\usepackage[pagebackref,breaklinks,colorlinks,allcolors=cvprblue]{hyperref}

\def\confName{CVPR}
\def\confYear{2025}

\title{Beyond Accuracy: Metrics that Uncover What Makes a `Good' Visual Descriptor}

\author{
Ethan Lin\textsuperscript{*1} \quad
Linxi Zhao\textsuperscript{1} \quad
Atharva Sehgal\textsuperscript{2} \quad
Jennifer J. Sun\textsuperscript{1} \\
\vspace{0.3em}
\textsuperscript{1}Cornell University \quad
\textsuperscript{2}University of Texas at Austin \\
\texttt{\{eyl45, lz586, jjs533\}@cornell.edu} \quad
\texttt{atharvas@utexas.edu}
}

\begin{document}
\maketitle

\begingroup
\renewcommand\thefootnote{}
\footnotetext{\textsuperscript{*}Correspondence to: \texttt{eyl45@cornell.edu}.}
\addtocounter{footnote}{-1}
\endgroup

\begingroup
\renewcommand{\thefootnote}{\fnsymbol{footnote}}
\footnotetext[2]{Code available at \url{https://github.com/ethan-y-lin/beyond_accuracy}.}
\endgroup

\begin{abstract}

Text-based visual descriptors—ranging from simple class names to more descriptive phrases—are widely used in visual concept discovery and image classification with vision-language models (VLMs). Their effectiveness, however, depends on a complex interplay of factors, including semantic clarity, presence in the VLM's pre-training data, and how well the descriptors serve as a meaningful representation space. In this work, we systematically analyze descriptor quality along two key dimensions: (1) representational capacity, and (2) relationship with VLM pre-training data. We evaluate a spectrum of descriptor generation methods, from zero-shot LLM-generated prompts to iteratively refined descriptors. Motivated by ideas from representation alignment and language understanding, we introduce two alignment-based metrics—Global Alignment and CLIP Similarity—that move beyond accuracy. These metrics shed light on how different descriptor generation strategies interact with foundation model properties, offering new ways to study descriptor effectiveness beyond accuracy evaluations.
\end{abstract}    
\section{Introduction}
\label{sec:intro}

Classification-by-description frameworks extend zero-shot classification by pairing each class with more informative text descriptors instead of a simple label. These descriptors—ranging from class names to full sentences—serve as queries to a vision-language model (VLM), which classifies images based on image similarity to each descriptor in a shared embedding space. Such methods have demonstrated improved classification performance and offer greater interpretability by making the model’s reasoning process more transparent.
\begin{figure}
    \centering
    \vspace{-2em}
    \includegraphics[width=1.0\linewidth]{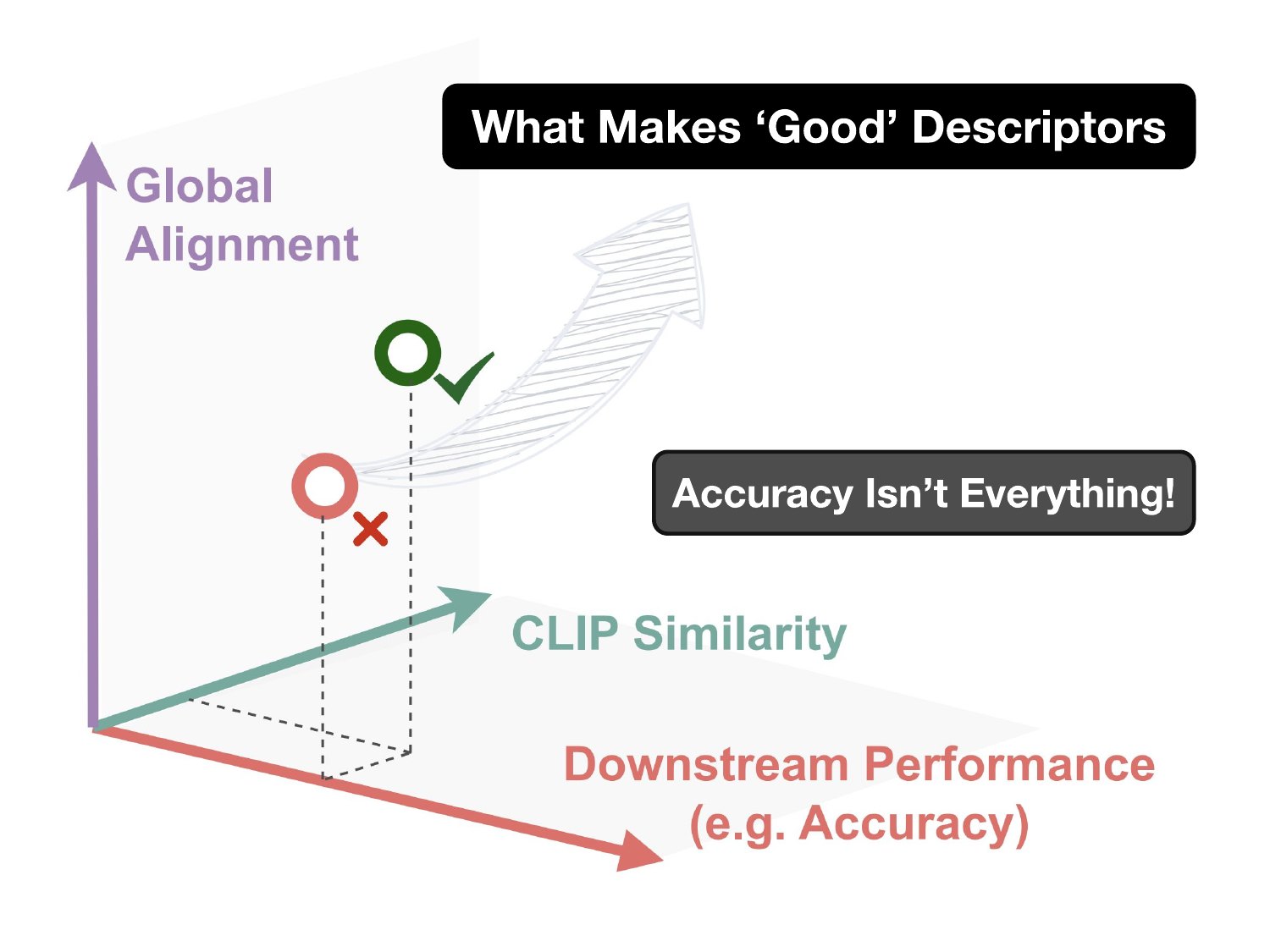}
    \caption{\textbf{Overview:} Achieving high accuracy does not guarantee that a set of descriptors is ``good''. Other factors like interpretability may suffer. Global alignment and CLIP similarity can serve as new metrics for evaluating and understanding different sets of visual descriptors. 
    }
    \label{fig:introduction}
\end{figure}

While these approaches have demonstrated success in image classification, a fundamental question remains: What constitutes a “good” textual visual descriptor? Descriptors can be obtained in many ways—from using class names directly~\cite{radford2021learning}, to leveraging large language models (LLMs) for zero-shot descriptions~\citep{menon2022visualclassificationdescriptionlarge}, to employing sophisticated iterative refinement algorithms~\cite{sehgal2025selfevolvingvisualconceptlibrary}. Descriptor quality is typically evaluated based on downstream classification accuracy. While informative, this metric offers limited insight into why certain descriptors succeed or fail, and provides little guidance for descriptor discovery~\citep{bhalla_interpreting_2024} or refinement beyond evolutionary search~\citep{llmmutate-24} or task-specific feedback loops~\citep{sehgal2025selfevolvingvisualconceptlibrary}. Moreover, higher accuracy does not necessarily imply more meaningful descriptors—prior work has shown that appending random words or characters to class names can sometimes improve performance~\cite{roth2023wafflingperformancevisualclassification}. Clearly, there is need for more principled methods to assess descriptor quality beyond accuracy.

In this work, we propose a novel approach to assessing descriptor quality by probing the relationship between textual descriptors and the underlying VLM. We focus on two key aspects: (1) the representational capacity of a set of descriptors—that is, how well they capture and reconstruct image semantics when used as a projection space; and (2) the relationship between a set of descriptors and CLIP’s pre-training data distribution. To study these factors systematically, we analyze a diverse range of text descriptor sets—spanning class names, zero-shot LLM-generated descriptions, and iteratively refined prompts—using two simple but informative metrics  (Figure \ref{fig:introduction}). Motivated by recent work on representational alignment~\citep{huh2024platonicrepresentationhypothesis}, we introduce a global alignment metric, \dinoalign{}, which measures the alignment between a vision model's image embedding space and the representation space induced by a set of descriptors using CLIP~\citep{radford_learning_2021}. We also propose \clipsim{}, inspired by findings on language model familiarity with training data~\citep{kandpal2023largelanguagemodelsstruggle}, which captures how descriptors relate to the CLIP training distribution. Together, these metrics move beyond accuracy to offer a deeper evaluation of text-based visual descriptor quality.

\vspace{1em} \noindent\textbf{Contributions.} This work makes these contributions:

\begin{itemize}

\item We propose two novel metrics — \textit{global representation alignment} and \textit{CLIP similarity} — to probe text descriptor quality beyond downstream accuracy.
\item We analyze how these metrics correlate with performance across fine-grained classification datasets, and present preliminary analyses that showcase how these metrics characterize the effectiveness of strong text-based visual descriptors.

\end{itemize}
\section{Related Work} \label{sec:related}

\paragraph{Concept Learning with VLMs.} Despite rapid improvements in large-scale visual pretraining \citep{radford_learning_2021, jia2021scaling, li2021align, mu2022slip, yaofilip2022, yu2022coca, alayrac_flamingo_2022, liu2024visual, liu2024improved}, vision-language models (VLMs) still struggle with even moderately complex perceptual questions \citep{tong_eyes_2024, kamath_whats_2023, alhamoud_vision-language_2025}. Concept learning has emerged as a promising direction, leveraging zero-shot queries to large language models (LLMs) to decompose complex perceptual reasoning tasks into logical combinations of simpler perceptual subtasks -- tasks that off-the-shelf VLMs can reliably solve \citep{menon_visual_2022, yan2023learning, yang2023languagebottlelanguagemodel, roth2023wafflingperformancevisualclassification}. These logical combinations are typically expressed as learnable linear combinations over an overcomplete set of concepts, sampled from an LLM. Prior work improves this initial concept set either by subsampling or by sampling new concepts using task-specific visual feedback from an LLM \citep{sehgal2025selfevolvingvisualconceptlibrary}. The quality of a set of concepts is quantitatively evaluated using downstream image classification performance. We complement prior work by analyzing the utility of a concept set through three orthogonal lenses: downstream concept accuracy, representational alignment, and alignment with VLM pre-training data. 
\paragraph{Representational Alignment with VLMs.} Representational alignment refers to the degree of similarity between the internal representations learned by different models \citep{sucholutsky2024gettingalignedrepresentationalalignment}. In the context of VLMs, research has studied both explicit training methods to maximize alignment \citep{jia2021scaling, radford_learning_2021} and the emergence of implicit alignment during training \citep{huh2024platonicrepresentationhypothesis,  tjandrasuwita2025understandingemergencemultimodalrepresentation}.

In this work, we provide the first study of such alignment metrics applied to representations induced by different sets of descriptors. We show how these alignment measures can serve as indicators of the downstream utility and meaningfulness of a descriptor set.

\section{Approach}

\begin{figure*}
    \centering
    \includegraphics[width=\linewidth]{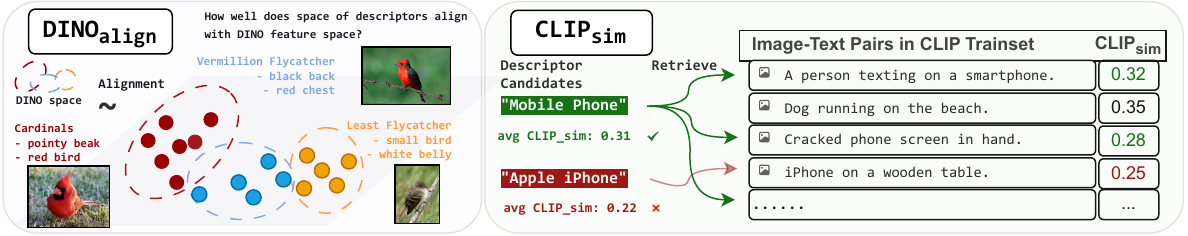}
\caption{\textbf{Overview of our descriptor evaluation metrics.} (\textbf{Left}) \dinoalign{} measures how well a set of descriptors preserves class structure by aligning the descriptor-induced similarity matrix with that of DINO image features. (\textbf{Right}) \clipsim{} assesses how well descriptors align with visual content by retrieving related image-text pairs from CLIP’s pre-training dataset and averaging their similarity scores.}
    \label{fig:method}
\end{figure*}

\subsection{Global Alignment Metric}
\label{sec:dino_align}
Classification by description using VLMs can be viewed as a semantic projection \cite{LM4CV}, where images are mapped to similarity scores over a set of textual descriptors. This forms a new representation space, with each dimension corresponding to a specific concept. We assess the quality of the descriptor-induced space using \dinoalign{}, a zero-shot metric that measures the correlation between image similarities in the descriptor space and a reference embedding space (Figure \ref{fig:method}). High alignment indicates that the descriptors capture meaningful visual structure.

Following work on the \textit{Platonic Representation Hypothesis}~\cite{huh2024platonicrepresentationhypothesis}, we use a \textbf{mutual $k$-nearest neighbor (M-KNN)} alignment metric to quantify structural similarity between two representations of the same data. Given representation matrices \( A \) and \( B \), we compute cosine-based $k$-nearest neighbor sets \( K_A \) and \( K_B \) and define the alignment score:
\[
\text{Alignment} = \frac{|KNN_A \cap KNN_B|}{k}
\]
This reflects the average neighborhood overlap between the two spaces. To evaluate the quality and interpretability of a set of text-based visual descriptors, we perform the following steps. 
\begin{enumerate}
    \item Remove class names from all descriptors to avoid bias.
    \item Combine descriptors across all classes to form a shared global descriptor set.
    \item Extract image embeddings \( X \) and concept embeddings \( Y \) using a vision-language model (VLM).
    \item Project images into the concept space via the image-concepts similarity matrix \( S = XY^\top \), where row \( i \) represents the image \( i \)'s similarity with each concept (columns).
    \item Compute DINO embeddings \( Z \) as the reference space and compute alignment using M-KNN.
\end{enumerate}

\subsection{CLIP Similarity Metric}
\label{sec:clip_sim}
Due to the imbalanced nature of CLIP’s training data, semantically similar text descriptors may not yield similar downstream performance. To analyze this, we propose \clipsim{}, a metric measuring how well descriptor candidates align with VLM pre-training data (Figure \ref{fig:method}).

We define two statistics per descriptor $d$, using its top-$k$ nearest captions $\{c_1, \dots, c_k\}$ retrieved by cosine similarity in CLIP’s text embedding space. Let $\text{sim}(a, b)$ denote the similarity between two L2-normalized embeddings following~\citet{kandpal2023largelanguagemodelsstruggle}. Exact or fuzzy string matching is avoided, as it often fails with fine-grained descriptors. We define two statistics per descriptor:

\begin{itemize}
    \item \textbf{Frequency}: number of captions above a similarity threshold $\tau$:
    \[
    \mathcal{I}_d = \left\{ i \in [k] \mid \text{sim}(d, c_i) > \tau \right\}, \quad
    \text{Freq}(d) = |\mathcal{I}_d|
    \]
    \item \textbf{Similarity}: average similarity between each matched caption $c_i$ and its paired image $I_i$:
    \[
    \text{Sim}(d) = \frac{1}{|\mathcal{I}_d|} \sum_{i \in \mathcal{I}_d} \text{sim}(c_i, I_i)
    \]
\end{itemize}

These two metrics captures how frequently a text descriptor appears in CLIP’s training set and how well it aligns with visual content in the training set, providing a proxy for how compatible it is with CLIP’s vision-language understanding.

Intuitively, high-quality descriptors should not only differentiate between image classes but also align with the inductive biases of vision-language models (VLMs) shaped by their training data. Such descriptors are more likely to yield better downstream performance.

Finally, the overall \clipsim{} score for a method is the average similarity across all descriptors:
\[
\clipsim{} = \frac{1}{|\mathcal{D}|} \sum_{d \in \mathcal{D}} \text{Sim}(d)
\]

\section{Experiments}

\subsection{Setup}
\paragraph{Dataset and Model}
We compute and analyze concept alignment and descriptor frequencies on three image classification benchmarks: CUB-200-2011~\cite{WahCUB_200_2011}, NABirds~\cite{Horn_2015_CVPR}, and CIFAR-100~\cite{krizhevsky2009learning}. All experiments are performed using the test sets of each dataset. These datasets are widely used by previous works and cover both fine-grained and general-purpose categories. We analyze CLIP’s pre-training data using the open-source DFN2B dataset.\footnote[1]{\url{huggingface.co/apple/DFN2B-CLIP-ViT-L-14}}
We use CLIP ViT-L/14~\cite{radford2021learning} to compute zero-shot accuracies and the image-concept similarity matrix. As the current SOTA image embedding space, we compute the reference image embedding using DinoV2-Base~\cite{oquab2024dinov2learningrobustvisual}.\footnote[2]{\url{huggingface.co/facebook/dinov2-base}}

\subsection{Descriptor Sets}
We evaluate several descriptor sets used for classification:
\begin{itemize}[leftmargin=1em, itemsep=0.2em]
    \item \textit{Class Name Prompt}: Uses the standard zero-shot format \emph{``An image of a \{\texttt{class name}\}''}, popularized by~\citet{radford2021learning}.
    
    \item \textit{CBD Concepts}: Descriptors from the original Classification by Description framework~\cite{menon2022visualclassificationdescriptionlarge}.
    
    \item \textit{ESCHER Iterative Descriptors}: Class-specific phrases refined through multiple ESCHER iterations~\cite{sehgal2025selfevolvingvisualconceptlibrary}.
    
    \item \textit{DCLIP (Randomized)}: Inspired by WaffleCLIP~\cite{roth2023wafflingperformancevisualclassification}; randomly samples a fixed pool of $N$ global descriptors. Each class is formatted as \emph{``\{\texttt{class name}\}, \{\texttt{descriptor\_i}\}''}.
    
    \item \textit{WaffleCLIP Randomization}: Forms descriptors by appending randomized tokens to general concepts and class names, e.g., \emph{``An image of a \{\texttt{concept}\}: \{\texttt{class name}\}, which has !32d, \#tjli, \^{ }fs0.''}
\end{itemize}

\subsection{Research Questions}
\label{sec:research_questions}

\begin{figure}[ht]
    \centering
    \begin{minipage}{\columnwidth}
        \centering
        \begin{subfigure}[t]{0.32\columnwidth}
            \centering
            \includegraphics[width=\linewidth]{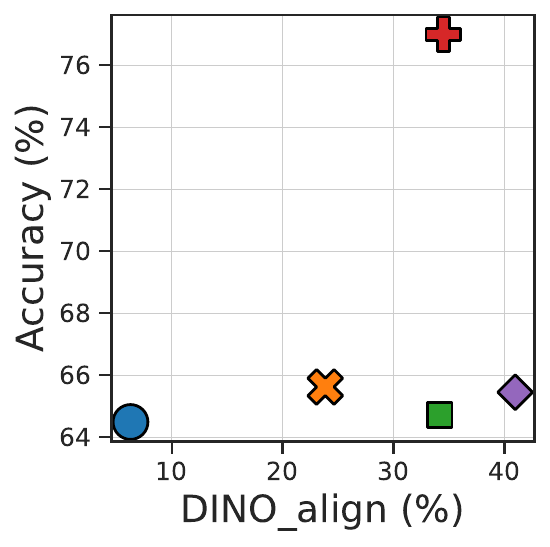}
            \caption{\textsc{CIFAR-100}}
        \end{subfigure}
        \hfill
        \begin{subfigure}[t]{0.32\columnwidth}
            \centering
            \includegraphics[width=\linewidth]{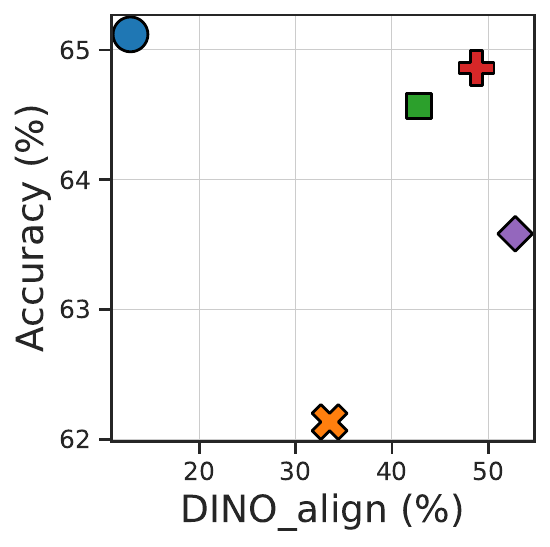}
            \caption{\textsc{CUB}}
        \end{subfigure}
        \hfill
        \begin{subfigure}[t]{0.32\columnwidth}
            \centering
            \includegraphics[width=\linewidth]{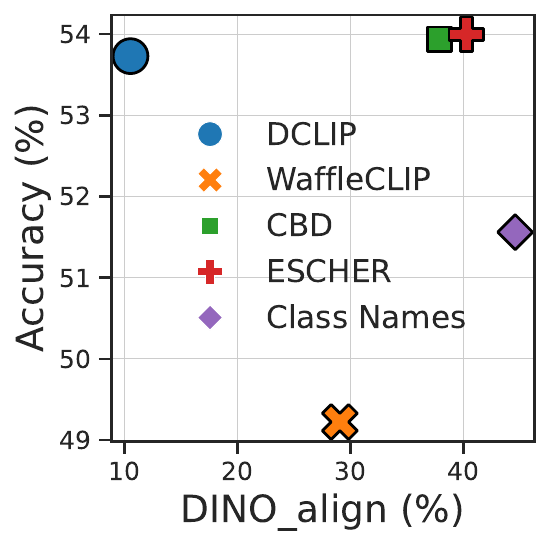}
            \caption{\textsc{NABirds}}
        \end{subfigure}
        \caption{\textbf{Classification accuracy versus alignment score} across all datasets. Alignment consistently reflects the quality of descriptor sets, regardless of their classification accuracy.}
        \label{fig:class_noclass_alignment_subfigs}
    \end{minipage}
\end{figure}

\paragraph{Can alignment metrics serve as reliable unsupervised indicators of descriptor set quality or refinement progress?}

Figure~\ref{fig:class_noclass_alignment_subfigs} compares accuracy and alignment across datasets. Since alignment is intended to reflect the representational quality of a descriptor set, we remove class names from each descriptor to isolate its effectiveness. The resulting scores follow a consistent pattern:

{\small
\begin{displaymath}
\textsc{DCLIP} < \textsc{WaffleCLIP} < \textsc{CBD} \leq \textsc{ESCHER} < \textsc{Class Name}
\end{displaymath}
}

Viewing classification by description as a form of semantic projection—analogous to PCA—the consistently high alignment scores of class names across datasets suggest that they function like the first principal components: basis directions that capture the highest variance and most distinguishing visual features. When a class-agnostic descriptor set achieves similar alignment, it likely reflects strong semantic grounding with respect to the image space. ESCHER achieves this more reliably than CBD, highlighting the effectiveness of its iterative refinement. In contrast, \textsc{DCLIP} and \textsc{WaffleCLIP} exhibit much lower alignment despite comparable accuracy, reinforcing that randomly generated descriptors lack meaningful structure. These results suggest that alignment is a consistent and informative signal for assessing semantic quality, even in the absence of class supervision.

\paragraph{Does an iterative algorithm like ESCHER converge toward descriptors that are naturally aligned with CLIP’s vision-language priors?}

\begin{figure} [th]
    \centering
    \includegraphics[width=\linewidth]{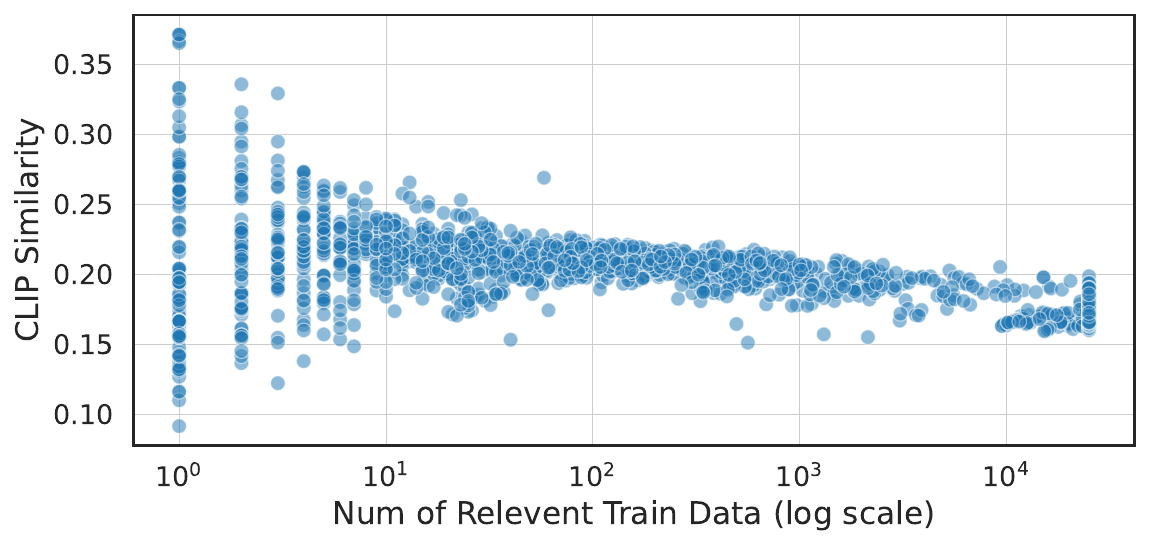}
    \caption{Distribution of descriptors by frequency and corresponding CLIP similarity on \textsc{NABirds}. The consistent negative trend suggests that low-frequency descriptors are often more visually grounded and better aligned with CLIP’s pretraining space.}

    \label{fig:freq}
\end{figure}

\begin{figure} [ht]
    \centering
    \begin{subfigure}[t]{0.32\columnwidth}
            \centering
            \includegraphics[width=\linewidth]{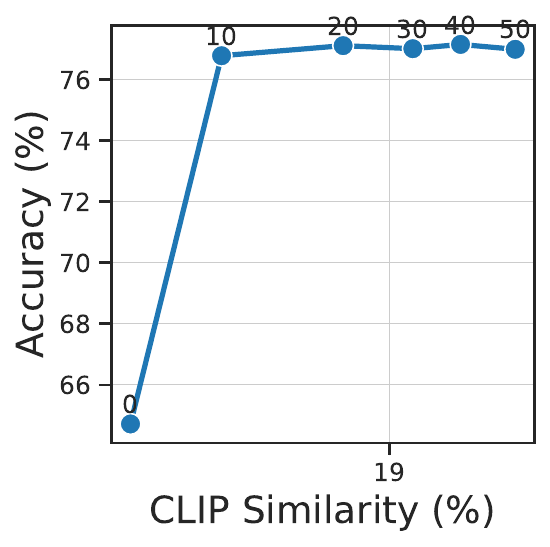}
            \caption{\textsc{CIFAR-100}}
        \end{subfigure}
        \hfill
        \begin{subfigure}[t]{0.32\columnwidth}
            \centering
            \includegraphics[width=\linewidth]{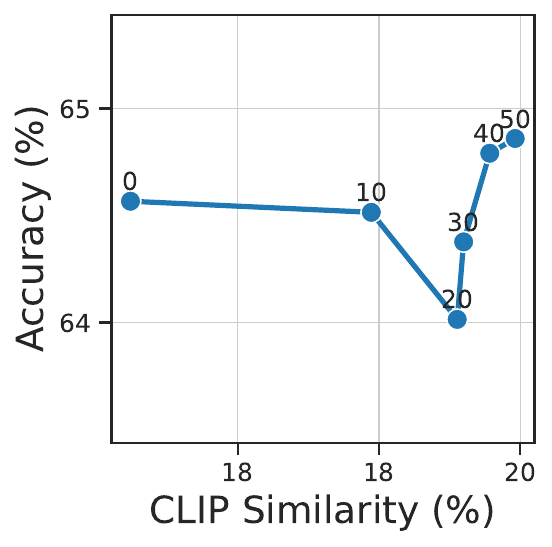}
            \caption{\textsc{CUB}}
        \end{subfigure}
        \hfill
        \begin{subfigure}[t]{0.32\columnwidth}
            \centering
            \includegraphics[width=\linewidth]{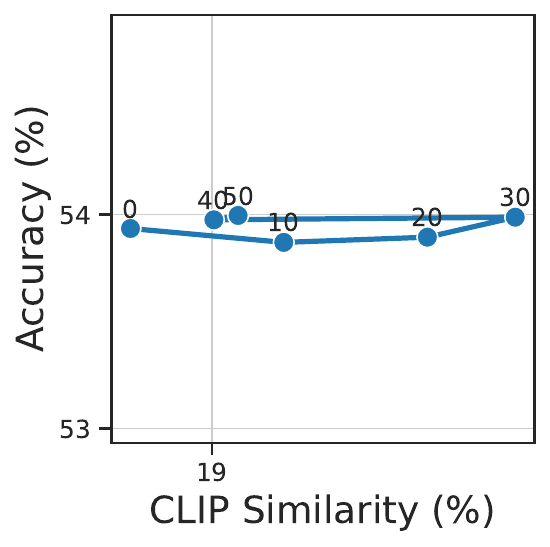}
            \caption{\textsc{NABirds}}
        \end{subfigure}
    \caption{We track how descriptor alignment with CLIP’s vision-language priors and classification accuracy evolve through iterative refinement.}
    \label{fig:iteration}
    \vspace{-0.2em}
\end{figure}

As ESCHER iteratively updates text descriptors for a downstream task, we expect the refined descriptors to not only become more discriminative but also better aligned with CLIP’s internal representation space. By tracking \clipsim{}, we can examine whether ESCHER learns to ``speak the language'' of CLIP’s training data.

To build intuition about CLIP’s priors, we first analyze the relationship between descriptor frequency and CLIP similarity. As shown in Figure~\ref{fig:freq}, we observe a consistent negative correlation between frequency and image-text similarity—a somewhat counterintuitive trend. One possible explanation is that frequent descriptors tend to be coarser and more ambiguous (e.g., ``animal''), making it harder for CLIP to associate them with a consistent visual pattern. In contrast, rarer descriptors often refer to more specific, visually grounded concepts that CLIP can align with more reliably.

Figure~\ref{fig:iteration} shows how descriptor quality evolves over ESCHER iterations. On CIFAR-100 and CUB, both \clipsim{} and classification accuracy improve steadily, indicating that descriptors become increasingly fine-grained and better aligned with CLIP’s representation space. On NABirds, the trend is less pronounced—accuracy remains largely stable across iterations, with a small dip at epoch 40 likely due to noise. These results support our hypothesis: ESCHER’s iterative refinement generally improves both CLIP alignment and downstream performance.
\section{Conclusion}
We present a systematic study of text-based visual descriptors, moving beyond accuracy to examine their alignment with visual features and CLIP’s pre-training distribution. Across a range of descriptor types, we find that alignment metrics often correlate with intuitive concept quality—even without access to labels. Notably, iteratively evolved descriptors become better aligned and yield improved downstream performance. These results underscore the value of alignment and frequency analysis as tools for understanding descriptor behavior and foundation model interactions, laying the groundwork for more principled descriptor evaluation and design in future work.

\section*{Acknowledgments}
This work is partially supported by LinkedIn-Cornell Partnership.

{
    \small
    \bibliographystyle{ieeenat_fullname}
    \bibliography{main}

\begin{thebibliography}{30}
\providecommand{\natexlab}[1]{#1}
\providecommand{\url}[1]{\texttt{#1}}
\expandafter\ifx\csname urlstyle\endcsname\relax
  \providecommand{\doi}[1]{doi: #1}\else
  \providecommand{\doi}{doi: \begingroup \urlstyle{rm}\Url}\fi

\bibitem[Alayrac et~al.(2022)Alayrac, Donahue, Luc, Miech, Barr, Hasson, Lenc, Mensch, Millican, Reynolds, Ring, Rutherford, Cabi, Han, Gong, Samangooei, Monteiro, Menick, Borgeaud, Brock, Nematzadeh, Sharifzadeh, Binkowski, Barreira, Vinyals, Zisserman, and Simonyan]{alayrac_flamingo_2022}
Jean-Baptiste Alayrac, Jeff Donahue, Pauline Luc, Antoine Miech, Iain Barr, Yana Hasson, Karel Lenc, Arthur Mensch, Katie Millican, Malcolm Reynolds, Roman Ring, Eliza Rutherford, Serkan Cabi, Tengda Han, Zhitao Gong, Sina Samangooei, Marianne Monteiro, Jacob Menick, Sebastian Borgeaud, Andrew Brock, Aida Nematzadeh, Sahand Sharifzadeh, Mikolaj Binkowski, Ricardo Barreira, Oriol Vinyals, Andrew Zisserman, and Karen Simonyan.
\newblock Flamingo: a {Visual} {Language} {Model} for {Few}-{Shot} {Learning}, 2022.
\newblock arXiv:2204.14198 [cs].

\bibitem[Alhamoud et~al.(2025)Alhamoud, Alshammari, Tian, Li, Torr, Kim, and Ghassemi]{alhamoud_vision-language_2025}
Kumail Alhamoud, Shaden Alshammari, Yonglong Tian, Guohao Li, Philip Torr, Yoon Kim, and Marzyeh Ghassemi.
\newblock Vision-language models do not understand negation, 2025.

\bibitem[Bhalla et~al.(2024)Bhalla, Oesterling, Srinivas, Calmon, and Lakkaraju]{bhalla_interpreting_2024}
Usha Bhalla, Alex Oesterling, Suraj Srinivas, Flavio~P. Calmon, and Himabindu Lakkaraju.
\newblock Interpreting {CLIP} with {Sparse} {Linear} {Concept} {Embeddings} ({SpLiCE}), 2024.
\newblock arXiv:2402.10376.

\bibitem[Chiquier et~al.(2024)Chiquier, Mall, and Vondrick]{llmmutate-24}
Mia Chiquier, Utkarsh Mall, and Carl Vondrick.
\newblock Evolving interpretable visual classifiers with large language models.
\newblock \emph{CoRR}, 2024.

\bibitem[Huh et~al.(2024)Huh, Cheung, Wang, and Isola]{huh2024platonicrepresentationhypothesis}
Minyoung Huh, Brian Cheung, Tongzhou Wang, and Phillip Isola.
\newblock The platonic representation hypothesis, 2024.

\bibitem[Jia et~al.(2021)Jia, Yang, Xia, Chen, Parekh, Pham, Le, Sung, Li, and Duerig]{jia2021scaling}
Chao Jia, Yinfei Yang, Ye Xia, Yi-Ting Chen, Zarana Parekh, Hieu Pham, Quoc Le, Yun-Hsuan Sung, Zhen Li, and Tom Duerig.
\newblock Scaling up visual and vision-language representation learning with noisy text supervision.
\newblock In \emph{International conference on machine learning}, pages 4904--4916. PMLR, 2021.

\bibitem[Kamath et~al.(2023)Kamath, Hessel, and Chang]{kamath_whats_2023}
Amita Kamath, Jack Hessel, and Kai-Wei Chang.
\newblock What's "up" with vision-language models? {Investigating} their struggle with spatial reasoning, 2023.
\newblock arXiv:2310.19785 [cs].

\bibitem[Kandpal et~al.(2023)Kandpal, Deng, Roberts, Wallace, and Raffel]{kandpal2023largelanguagemodelsstruggle}
Nikhil Kandpal, Haikang Deng, Adam Roberts, Eric Wallace, and Colin Raffel.
\newblock Large language models struggle to learn long-tail knowledge, 2023.

\bibitem[Krizhevsky et~al.(2009)Krizhevsky, Nair, and Hinton]{krizhevsky2009learning}
Alex Krizhevsky, Vinod Nair, and Geoffrey Hinton.
\newblock Cifar-100 (canadian institute for advanced research).
\newblock 2009.

\bibitem[Li et~al.(2021)Li, Selvaraju, Gotmare, Joty, Xiong, and Hoi]{li2021align}
Junnan Li, Ramprasaath Selvaraju, Akhilesh Gotmare, Shafiq Joty, Caiming Xiong, and Steven Chu~Hong Hoi.
\newblock Align before fuse: Vision and language representation learning with momentum distillation.
\newblock \emph{Advances in neural information processing systems}, 34:\penalty0 9694--9705, 2021.

\bibitem[Liu et~al.(2024{\natexlab{a}})Liu, Li, Li, and Lee]{liu2024improved}
Haotian Liu, Chunyuan Li, Yuheng Li, and Yong~Jae Lee.
\newblock Improved baselines with visual instruction tuning.
\newblock In \emph{Proceedings of the IEEE/CVF Conference on Computer Vision and Pattern Recognition}, pages 26296--26306, 2024{\natexlab{a}}.

\bibitem[Liu et~al.(2024{\natexlab{b}})Liu, Li, Wu, and Lee]{liu2024visual}
Haotian Liu, Chunyuan Li, Qingyang Wu, and Yong~Jae Lee.
\newblock Visual instruction tuning.
\newblock \emph{Advances in neural information processing systems}, 36, 2024{\natexlab{b}}.

\bibitem[Menon and Vondrick(2022{\natexlab{a}})]{menon2022visualclassificationdescriptionlarge}
Sachit Menon and Carl Vondrick.
\newblock Visual classification via description from large language models, 2022{\natexlab{a}}.

\bibitem[Menon and Vondrick(2022{\natexlab{b}})]{menon_visual_2022}
Sachit Menon and Carl Vondrick.
\newblock Visual {Classification} via {Description} from {Large} {Language} {Models}.
\newblock 2022{\natexlab{b}}.

\bibitem[Mu et~al.(2022)Mu, Kirillov, Wagner, and Xie]{mu2022slip}
Norman Mu, Alexander Kirillov, David Wagner, and Saining Xie.
\newblock Slip: Self-supervision meets language-image pre-training.
\newblock In \emph{European conference on computer vision}, pages 529--544. Springer, 2022.

\bibitem[Oquab et~al.(2024)Oquab, Darcet, Moutakanni, Vo, Szafraniec, Khalidov, Fernandez, Haziza, Massa, El-Nouby, Assran, Ballas, Galuba, Howes, Huang, Li, Misra, Rabbat, Sharma, Synnaeve, Xu, Jegou, Mairal, Labatut, Joulin, and Bojanowski]{oquab2024dinov2learningrobustvisual}
Maxime Oquab, Timothée Darcet, Théo Moutakanni, Huy Vo, Marc Szafraniec, Vasil Khalidov, Pierre Fernandez, Daniel Haziza, Francisco Massa, Alaaeldin El-Nouby, Mahmoud Assran, Nicolas Ballas, Wojciech Galuba, Russell Howes, Po-Yao Huang, Shang-Wen Li, Ishan Misra, Michael Rabbat, Vasu Sharma, Gabriel Synnaeve, Hu Xu, Hervé Jegou, Julien Mairal, Patrick Labatut, Armand Joulin, and Piotr Bojanowski.
\newblock Dinov2: Learning robust visual features without supervision, 2024.

\bibitem[Radford et~al.(2021{\natexlab{a}})Radford, Kim, Hallacy, Ramesh, Goh, Agarwal, Sastry, Askell, Mishkin, Clark, Krueger, and Sutskever]{radford_learning_2021}
Alec Radford, Jong~Wook Kim, Chris Hallacy, Aditya Ramesh, Gabriel Goh, Sandhini Agarwal, Girish Sastry, Amanda Askell, Pamela Mishkin, Jack Clark, Gretchen Krueger, and Ilya Sutskever.
\newblock Learning {Transferable} {Visual} {Models} {From} {Natural} {Language} {Supervision}, 2021{\natexlab{a}}.
\newblock Number: arXiv:2103.00020 arXiv:2103.00020 [cs].

\bibitem[Radford et~al.(2021{\natexlab{b}})Radford, Kim, Hallacy, Ramesh, Goh, Agarwal, Sastry, Askell, Mishkin, Clark, and {others}]{radford2021learning}
Alec Radford, Jong~Wook Kim, Chris Hallacy, Aditya Ramesh, Gabriel Goh, Sandhini Agarwal, Girish Sastry, Amanda Askell, Pamela Mishkin, Jack Clark, and {others}.
\newblock Learning transferable visual models from natural language supervision.
\newblock In \emph{International conference on machine learning}, pages 8748--8763. PMLR, 2021{\natexlab{b}}.

\bibitem[Roth et~al.(2023)Roth, Kim, Koepke, Vinyals, Schmid, and Akata]{roth2023wafflingperformancevisualclassification}
Karsten Roth, Jae~Myung Kim, A.~Sophia Koepke, Oriol Vinyals, Cordelia Schmid, and Zeynep Akata.
\newblock Waffling around for performance: Visual classification with random words and broad concepts, 2023.

\bibitem[Sehgal et~al.(2025)Sehgal, Yuan, Hu, Yue, Sun, and Chaudhuri]{sehgal2025selfevolvingvisualconceptlibrary}
Atharva Sehgal, Patrick Yuan, Ziniu Hu, Yisong Yue, Jennifer~J. Sun, and Swarat Chaudhuri.
\newblock Self-evolving visual concept library using vision-language critics, 2025.

\bibitem[Sucholutsky et~al.(2024)Sucholutsky, Muttenthaler, Weller, Peng, Bobu, Kim, Love, Cueva, Grant, Groen, Achterberg, Tenenbaum, Collins, Hermann, Oktar, Greff, Hebart, Cloos, Kriegeskorte, Jacoby, Zhang, Marjieh, Geirhos, Chen, Kornblith, Rane, Konkle, O'Connell, Unterthiner, Lampinen, Müller, Toneva, and Griffiths]{sucholutsky2024gettingalignedrepresentationalalignment}
Ilia Sucholutsky, Lukas Muttenthaler, Adrian Weller, Andi Peng, Andreea Bobu, Been Kim, Bradley~C. Love, Christopher~J. Cueva, Erin Grant, Iris Groen, Jascha Achterberg, Joshua~B. Tenenbaum, Katherine~M. Collins, Katherine~L. Hermann, Kerem Oktar, Klaus Greff, Martin~N. Hebart, Nathan Cloos, Nikolaus Kriegeskorte, Nori Jacoby, Qiuyi Zhang, Raja Marjieh, Robert Geirhos, Sherol Chen, Simon Kornblith, Sunayana Rane, Talia Konkle, Thomas~P. O'Connell, Thomas Unterthiner, Andrew~K. Lampinen, Klaus-Robert Müller, Mariya Toneva, and Thomas~L. Griffiths.
\newblock Getting aligned on representational alignment, 2024.

\bibitem[Tjandrasuwita et~al.(2025)Tjandrasuwita, Ekbote, Ziyin, and Liang]{tjandrasuwita2025understandingemergencemultimodalrepresentation}
Megan Tjandrasuwita, Chanakya Ekbote, Liu Ziyin, and Paul~Pu Liang.
\newblock Understanding the emergence of multimodal representation alignment, 2025.

\bibitem[Tong et~al.(2024)Tong, Liu, Zhai, Ma, LeCun, and Xie]{tong_eyes_2024}
Shengbang Tong, Zhuang Liu, Yuexiang Zhai, Yi Ma, Yann LeCun, and Saining Xie.
\newblock Eyes {Wide} {Shut}? {Exploring} the {Visual} {Shortcomings} of {Multimodal} {LLMs}.
\newblock In \emph{2024 {IEEE}/{CVF} {Conference} on {Computer} {Vision} and {Pattern} {Recognition} ({CVPR})}, pages 9568--9578, Seattle, WA, USA, 2024. IEEE.

\bibitem[Van~Horn et~al.(2015)Van~Horn, Branson, Farrell, Haber, Barry, Ipeirotis, Perona, and Belongie]{Horn_2015_CVPR}
Grant Van~Horn, Steve Branson, Ryan Farrell, Scott Haber, Jessie Barry, Panos Ipeirotis, Pietro Perona, and Serge Belongie.
\newblock Building a bird recognition app and large scale dataset with citizen scientists: The fine print in fine-grained dataset collection.
\newblock In \emph{Proceedings of the IEEE Conference on Computer Vision and Pattern Recognition (CVPR)}, 2015.

\bibitem[Wah et~al.(2011)Wah, Branson, Welinder, Perona, and Belongie]{WahCUB_200_2011}
C. Wah, S. Branson, P. Welinder, P. Perona, and S. Belongie.
\newblock Caltech-ucsd birds-200-2011.
\newblock Technical Report CNS-TR-2011-001, California Institute of Technology, 2011.

\bibitem[Yan et~al.(2023{\natexlab{a}})Yan, Wang, Zhong, Dong, He, Lu, Wang, Shang, and McAuley]{LM4CV}
An Yan, Yu Wang, Yiwu Zhong, Chengyu Dong, Zexue He, Yujie Lu, William Wang, Jingbo Shang, and Julian~J. McAuley.
\newblock Learning concise and descriptive attributes for visual recognition.
\newblock \emph{CoRR}, abs/2308.03685, 2023{\natexlab{a}}.

\bibitem[Yan et~al.(2023{\natexlab{b}})Yan, Wang, Zhong, Dong, He, Lu, Wang, Shang, and McAuley]{yan2023learning}
An Yan, Yu Wang, Yiwu Zhong, Chengyu Dong, Zexue He, Yujie Lu, William~Yang Wang, Jingbo Shang, and Julian McAuley.
\newblock Learning concise and descriptive attributes for visual recognition.
\newblock In \emph{Proceedings of the IEEE/CVF International Conference on Computer Vision}, pages 3090--3100, 2023{\natexlab{b}}.

\bibitem[Yang et~al.(2023)Yang, Panagopoulou, Zhou, Jin, Callison-Burch, and Yatskar]{yang2023languagebottlelanguagemodel}
Yue Yang, Artemis Panagopoulou, Shenghao Zhou, Daniel Jin, Chris Callison-Burch, and Mark Yatskar.
\newblock Language in a bottle: Language model guided concept bottlenecks for interpretable image classification, 2023.

\bibitem[Yao et~al.(2022)Yao, Huang, Hou, Lu, Niu, Xu, Liang, Li, Jiang, and Xu]{yaofilip2022}
Lewei Yao, Runhui Huang, Lu Hou, Guansong Lu, Minzhe Niu, Hang Xu, Xiaodan Liang, Zhenguo Li, Xin Jiang, and Chunjing Xu.
\newblock Filip: Fine-grained interactive language-image pre-training.
\newblock In \emph{International Conference on Learning Representations}, 2022.

\bibitem[Yu et~al.(2022)Yu, Wang, Vasudevan, Yeung, Seyedhosseini, and Wu]{yu2022coca}
Jiahui Yu, Zirui Wang, Vijay Vasudevan, Legg Yeung, Mojtaba Seyedhosseini, and Yonghui Wu.
\newblock Coca: Contrastive captioners are image-text foundation models.
\newblock \emph{Transactions on Machine Learning Research}, 2022.

\end{thebibliography}
}

\clearpage
\appendix
\renewcommand{\thesection}{\Alph{section}}
\setcounter{page}{1}
\maketitlesupplementary
\section{Experiment Setup}
\paragraph{Global Alignment (\dinoalign{}) Details} 
As defined in Section ~\ref{sec:dino_align}, \dinoalign{} quantifies the alignment between the image embedding space of DINOv2 and the representation space induced by a set of descriptors using the \textbf{Mutual-KNN} metric. The value of \( k \) is chosen based on the average number of images per class of a dataset. This is a simple yet effective heuristic as it generally captures the point where the alignment score stops increasing significantly. Additionally, when computing alignment on a set of descriptors, we remove class names from the descriptors to avoid bias. This is described in more detail in section \ref{sec:additional_analysis}.

\paragraph{CLIP Similarity (\clipsim{})}
As defined in Section~\ref{sec:clip_sim}, this metric quantifies how well a descriptor aligns with CLIP’s pretraining distribution. We set the similarity threshold $\tau = 0.7$ and compute statistics over the top 5\% of most similar caption-image pairs. To approximate CLIP’s training distribution, we randomly sample 5 million image-caption pairs from the DFN2B dataset.\footnote[1]{\url{https://huggingface.co/apple/DFN2B-CLIP-ViT-L-14}} Cosine similarity is used for both text-to-text and text-to-image comparisons.

\paragraph{Classification Accuracy} 
We evaluate zero-shot classification performance of a set of descriptors on a dataset using the original classification by description evaluation method ~\citep{menon_visual_2022}.

\paragraph{Descriptor Evolution Analysis}
To study how ESCHER refines descriptors over time, we evaluate two metrics at regular training checkpoints: CLIP alignment (\clipsim{}) and zero-shot classification accuracy. All experiments are conducted using the official ESCHER implementation \cite{sehgal2025selfevolvingvisualconceptlibrary}.

We perform this analysis on three datasets: CIFAR-100, CUB, and NABirds. For each dataset, we report both metrics at every 10 steps to track the evolution of descriptor quality throughout ESCHER’s iterative refinement process.

\section{Detailed Results}
\paragraph{CLIP similarity across descriptor sets}

\begin{figure}[ht]
    \centering
    \begin{minipage}{\columnwidth}
        \centering
        \begin{subfigure}[t]{0.32\columnwidth}
            \centering
            \includegraphics[width=\linewidth]{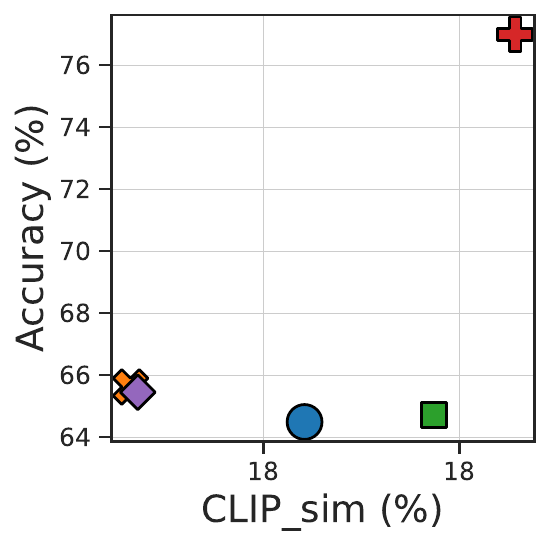}
            \caption{\textsc{CIFAR-100}}
        \end{subfigure}
        \hfill
        \begin{subfigure}[t]{0.32\columnwidth}
            \centering
            \includegraphics[width=\linewidth]{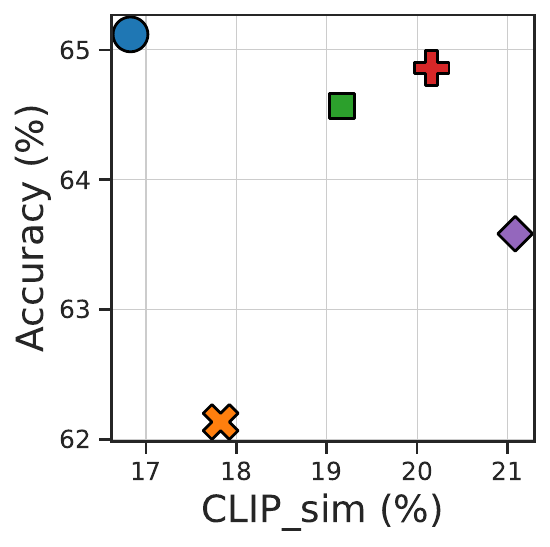}
            \caption{\textsc{CUB}}
        \end{subfigure}
        \hfill
        \begin{subfigure}[t]{0.32\columnwidth}
            \centering
            \includegraphics[width=\linewidth]{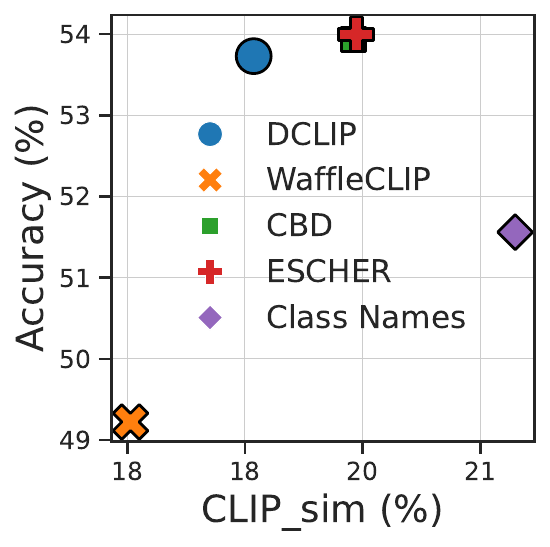}
            \caption{\textsc{NABirds}}
        \end{subfigure}
        \caption{\textbf{Classification accuracy versus \clipsim{}} across all datasets.}
        \label{fig:sim_subfigs}
    \end{minipage}
\end{figure}

Figure~\ref{fig:sim_subfigs} compares \clipsim{} across descriptor sets produced by different methods. The scores follow a consistent pattern:

{\scriptsize
\begin{displaymath}
\textsc{DCLIP}, \textsc{WaffleCLIP} < \textsc{CBD} \leq \textsc{ESCHER}
\end{displaymath}
}

This trend confirms that \clipsim{} provides a useful proxy for descriptor quality: descriptors that are more aligned with CLIP’s vision-language priors tend to show higher image-text similarity and better downstream performance. While most methods follow this pattern, we observe inconsistent results for \textsc{Class Name}, likely due to their coarse granularity and limited visual specificity.

\paragraph{Descriptor frequency vs. CLIP similarity}

\begin{figure} [ht]
    \centering
    \includegraphics[width=\linewidth]{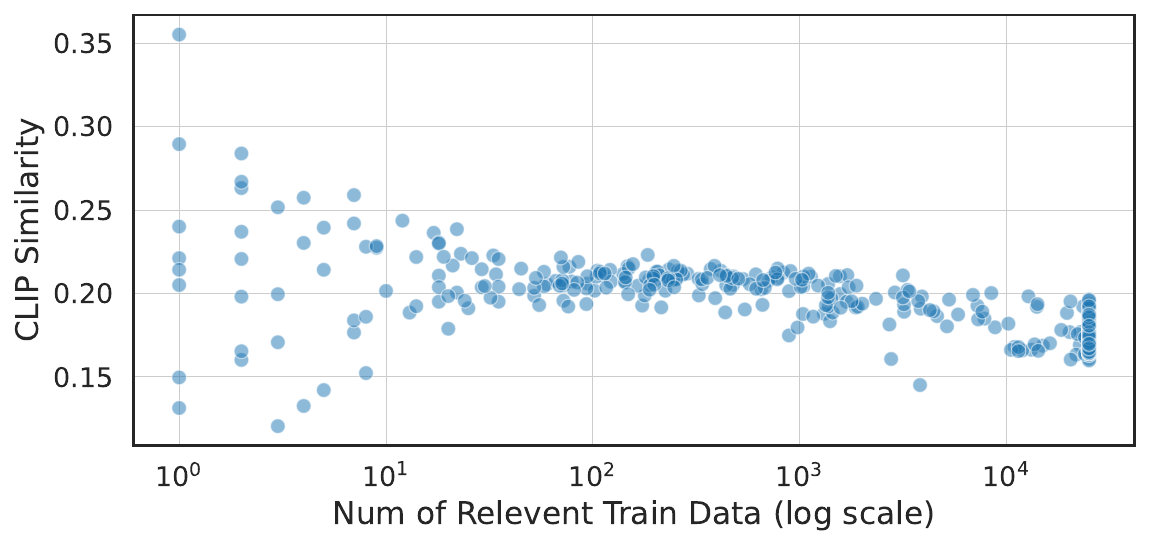}
    \caption{Distribution of descriptors by frequency (number of matching training captions) and corresponding CLIP similarity on \textsc{CIFAR-100}}
    \label{fig:freq_cifar}
\end{figure}

\begin{figure} [ht]
    \centering
    \includegraphics[width=\linewidth]{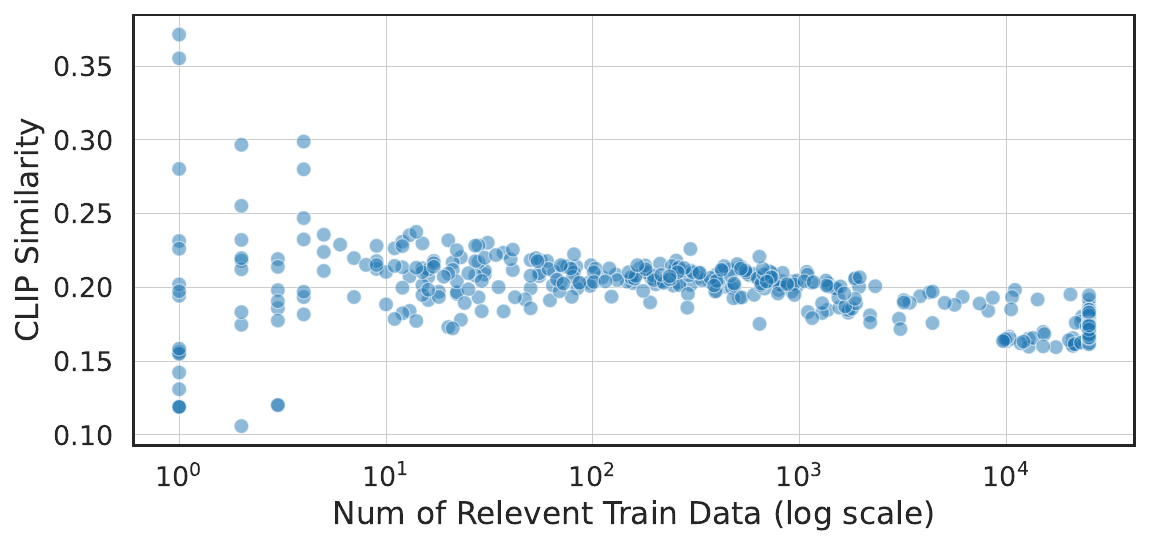}
    \caption{Distribution of descriptors by frequency (number of matching training captions) and corresponding CLIP similarity on \textsc{CUB}}
    \label{fig:freq_cub}
\end{figure}

To further probe how descriptors align with CLIP’s training distribution, we analyze the relationship between descriptor frequency (i.e., the number of matching training captions) and CLIP similarity. As shown in Figure~\ref{fig:freq_cub} and Figure~\ref{fig:freq_cifar}, we find a consistent negative correlation across datasets: lower-frequency descriptors tend to achieve higher CLIP similarity. This suggests that rare descriptors may capture more fine-grained, visually grounded concepts that are easier for CLIP to align with, despite being less common in the training set.


\section{Additional Analysis}
\label{sec:additional_analysis}

\paragraph{Class Name Bias in Alignment}

In many classification-by-description frameworks—including all those evaluated in this work—class names are prepended to each descriptor in the set. For example, descriptors associated with the class \textit{laysan albatross} are formatted as: ``laysan albatross, which is a \_\_\_\_''. While this improves performance, it also introduces a strong bias that makes it difficult to isolate the quality of the descriptors themselves.

\begin{figure}[ht]
    \centering
    \begin{minipage}{\columnwidth}
        \centering
        \begin{subfigure}[t]{0.32\columnwidth}
            \centering
            \includegraphics[width=\linewidth]{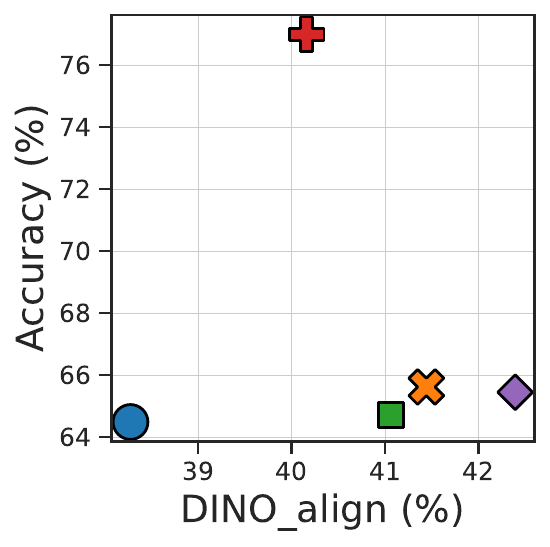}
            \caption{\textsc{CIFAR-100}}
        \end{subfigure}
        \hfill
        \begin{subfigure}[t]{0.32\columnwidth}
            \centering
            \includegraphics[width=\linewidth]{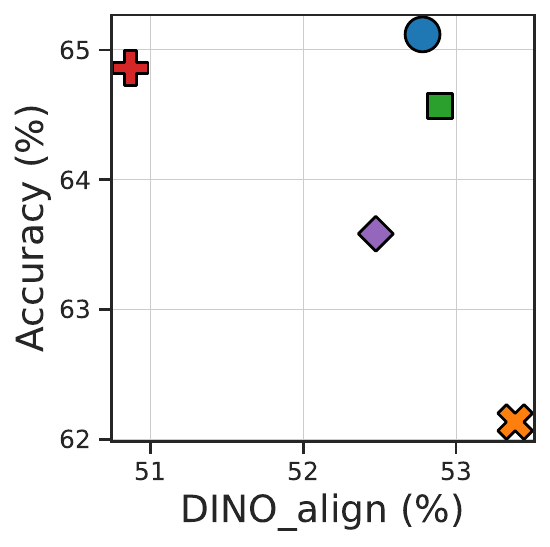}
            \caption{\textsc{CUB}}
        \end{subfigure}
        \hfill
        \begin{subfigure}[t]{0.32\columnwidth}
            \centering
            \includegraphics[width=\linewidth]{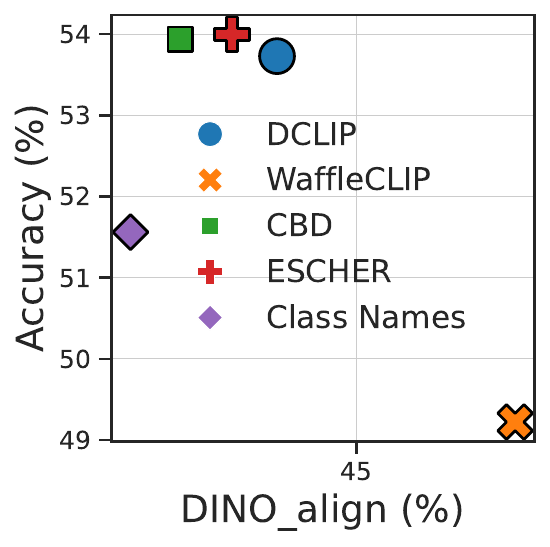}
            \caption{\textsc{NABirds}}
        \end{subfigure}
        \caption{\textbf{Classification accuracy versus alignment score} across all datasets wihtout removing class names from descriptors.}
        \label{fig:class_class_alignment_subfigs}
    \end{minipage}
\end{figure}

Figure~\ref{fig:class_class_alignment_subfigs} presents alignment and accuracy scores when class names are retained in all descriptors. In contrast to the clearer trends observed when class names are removed, no consistent patterns emerge across methods or datasets—highlighting how strongly class name inclusion can skew alignment metrics. For randomized methods such as \textsc{DCLIP} and \textsc{WaffleCLIP}, this dependence is expected, as their discriminative ability largely comes from the class name itself. However, the lack of structure in this setting also suggests that even more principled descriptor generation methods may be overly reliant on class name activation, rather than the intrinsic semantic quality of the descriptors.

\section{Limitations and Future Work}

This work presents a preliminary investigation into two novel alignment-based metrics for evaluating the quality of descriptor sets. While we demonstrate their utility across a selected set of descriptor generation methods, our analysis is not exhaustive and opens several avenues for future exploration.

Both metrics have limitations. For \textbf{\dinoalign{}}, computational costs scale with both the number of descriptors and the number of samples, limiting its feasibility for large-scale settings. Moreover, the interpretability of the KNN-based alignment metric likely diminishes when applied to large descriptor sets, and its sensitivity to class imbalance can make cross-dataset comparisons unreliable. As such, alignment is likely most informative when used on compact descriptor sets (e.g., fewer than 100 descriptors), particularly in few-shot settings such as those explored in LM4CV \cite{LM4CV} or in evolutionary descriptor search \cite{llmmutate-24}, where performance is achieved without reliance on class names.

For \textbf{\clipsim{}}, our current analysis is based on a 5M subset of the pre-training data. While this provides a useful approximation, the results could be made more robust by scaling up to a larger portion of the dataset. Future work should explore how descriptor similarity trends generalize across broader samples of the training distribution.

Additionally, our evaluation focuses on a subset of descriptor generation methods. Future studies should incorporate a wider range of approaches, including those that explicitly avoid class name dependence, such as LM4CV \cite{LM4CV}, or evolutionary search strategies \cite{llmmutate-24}. Applying our metrics to these methods would offer further insight into the quality of different sets of descriptors.

Overall, our metrics lay the groundwork for deeper, label-free evaluations of text-based visual descriptors, and we believe they can help inform the development of more interpretable and effective representations in vision-language models using these descriptors.

\end{document}